%% file: main.tex
\pdfoutput=1

\documentclass[11pt]{article}

\usepackage[]{acl}

\usepackage{times}
\usepackage{latexsym}
\usepackage[T1]{fontenc}

\usepackage[utf8]{inputenc}

\usepackage{microtype}
\usepackage{graphicx}
\usepackage{multirow}
\usepackage{stfloats}
\usepackage{pdfpages}

\title{Legal Prompting: Teaching a Language Model to Think Like a Lawyer}

\author{Fangyi Yu\Thanks{ The first author was a student at Ontario Tech University at the time of submission of this paper. This is work done during an internship at Thomson Reuters Labs. Please reach out to \textit{fangyi.yu@ontariotechu.net} for questions.}\\
   Thomson Reuters Labs \\
  19 Duncan Street \\
  Toronto, ON M5H 3G6 \\
  Canada \\
  \texttt{fangyi.yu@tr.com} \\\And
  Lee Quartey \\
  Thomson Reuters Labs \\
  3 Times Square \\
  New York, NY 10036\\
  United States \\
  \texttt{lee.quartey@tr.com} \\\And
  Frank Schilder \\
  Thomson Reuters Labs \\
  610 Opperman Drive\\
  Eagan, MN 55123\\
  United States \\
   \texttt{frank.schilder@tr.com} 
   }

\begin{document}
\maketitle
\begin{abstract}
Large language models that are capable of zero or few-shot prompting approaches have given rise to the new research area of prompt engineering.
Recent advances showed that for example Chain-of-Thought (CoT) prompts can improve arithmetic or common sense tasks significantly.
We explore how such approaches fare with legal reasoning tasks and take the COLIEE entailment task based on the Japanese Bar exam for testing zero-shot/few-shot and fine-tuning approaches.
Our findings show that while CoT prompting and fine-tuning with explanations approaches show improvements, the best results are produced by prompts that are derived from specific legal reasoning techniques such as  IRAC (Issue, Rule, Application, Conclusion).
Based on our experiments we improve the 2021 best result from 0.7037 accuracy to 0.8148 accuracy and beat the 2022 best system of 0.6789 accuracy with an accuracy of 0.7431.
\end{abstract}

\input{introduction}

\input{legal_entailment}

\input{previous_approaches}

\input{experiments}
 
\input{conclusions}

\bibliography{references}

\appendix

\section{Appendix}
\input{appendix}

\end{document}

%% file: introduction.tex
\section{Introduction}
One of the most fundamental responsibilities of a legal professional is to leverage analytical problem solving skills in applying a seemingly endless universe of law to an equally endless universe of scenarios. 
Such laws can be confusing and contradictory, well-reasoned rationales may be far from straightforward, and application may be inconsistent. 
Accordingly, the passing rates for the bar examination -- a critical step toward becoming a practicing attorney in many countries -- range from about 80\% in the United States in 2021\footnote{https://www.reuters.com/legal/legalindustry/bar-exam-pass-rate-dropped-last-year-first-time-testers-2022-04-26/} to 39.2\% in Japan in 2020 (widely regarded as one of the most difficult of all bar examinations)\footnote{https://www.nippon.com/en/japan-data/h00942/}.
The best attorneys are not only able to achieve these tasks with ease, but they are also able to effectively \textit{explain} the basis for their work. 
More often than not, a simple binary response to a given legal question is far from acceptable without the appropriate rationale to conclude it, perhaps accompanied by a chain of references to applicable statutes or a legal reasoning technique such as \textit{Issue, Rule, Application, Conclusion} \cite{burton2017think}.

Application of reason-based prompting mechanisms with large language models is becoming an increasingly prevalent focus area in natural language processing research.  
Models like Open\-AI's GPT-3 achieve satisfactory results -- 81\% accuracy with a zero-shot approach and 82.8\% accuracy with a few-shot approach -- on commonsense reasoning tasks drawn from the PhysicalQA dataset \cite{brown2020language}, but, as we observe, struggle significantly with more specialized domain data.
Further research builds on these basic queries by implementing a variety of so-called prompt engineering approaches, which range from soliciting a model to ``think step by step'' in producing incremental reasoning to underscore a given response \cite{kojima2022large}, to leveraging an iterative cycle of model generated rationales to bootstrap its own ability to produce more elaborate reasoning approaches \cite{zelikman2022star}; these approaches have demonstrated a nomimal improvement in a language model's ability to rationalize a correct response to a particular baseline query.

Our research aims to explore the effects of such approaches on highly specialized domain data, namely that in the legal field drawn from the Japanese bar exam. 
We frame our approach around the University of Alberta's annual Competition on Legal Information Extraction/Entailment -- or COLIEE -- event \cite{Rabelo_2022}, in which certain subtasks are devoted to reasoning through legal hypotheses given contextual articles; the COLIEE data itself solicits a yes/no response following a proposed hypothesis and supporting legal statutes. 

We first explore zero through few-shot approaches using pretrained LLMs, coupled with prompts either drawn from existing work (e.g., ``Let's think step by step'' \cite{kojima2022large}) or generated by ourselves (e.g., ``Please determine if the following hypothesis is True or False based on the given premise''). 

Additionally, we assess the impacts of fine tuning an LLM to infer binary responses both with and without explanations (either machine generated or extracted from the supporting premise). 
The best results on the COLIEE 2021 test set, though, result from a zero-shot, legal-prompt approach, which surpasses state of the art COLIEE performance by 15.79\%. 
Similarly, an 8-shot approach yields the best performance on 2022 COLIEE test data, with an overall accuracy improvement of 9.46\%.

Our experiments show that few-shot and fine-tuning with explanation approaches show good and consistent results for the two test sets of the COLIEE competition we used for evaluation. Zero-shot and a fine-tuning approach using the labels only show more inconsistent results across the two years. The zero-shot with legal reasoning approach shows the best result for one year only and may be more prone to overfitting to a specific test set indicating that further research of those prompting approaches is needed.

%% file: legal_entailment.tex
\section{Legal Entailment task}

The COLIEE competition \cite{Rabelo_2022} has been carried out since 2014 driving research in the area of legal retrieval and entailment. 
Two of the competition's tasks are using data from the Japanese bar exam.
The exam requires lawyers to determine whether a given legal statement is true or false.
In order to answer the questions, the lawyer has to first determine which articles of the Japanese 
statutes are most relevant to the given question. Task 3 of the competition covers this retrieval task.
Given one or more articles relevant to the question, the lawyer has then to determine whether the 
question (i.e., the hypothesis) is true or false given the selected articles (i.e., the premise).
This task is captured as task 4 in the COLIEE competition and we are focusing on this task with our work on legal reasoning:

\begin{description}
\item[Hypothesis:]  If the grounds of commencement of assistance cease to exist, the family court may rescind the decision for commencement of assistance without any party's request.
\item[Premise:] Article 18 (1) If the grounds prescribed in the main clause of Article 15, paragraph (1) cease to exist, the family court must rescind the decision for commencement of assistance at the request of the person in question, that person's spouse, that person's relative within the fourth degree of kinship, the guardian of a minor, the supervisor of a minor's guardian, the assistant, the assistant's supervisor, or a public prosecutor.\\
(2) At the request of a person as prescribed in the preceding paragraph, the family court may rescind all or part of the decision referred to in paragraph (1) of the preceding Article.
\item[Entailment:] NO
\end{description}

More formally the task is defined by the organizers as a legal entailment task:

Given a question $Q$, after retrieving relevant articles $S_1$, $S_2$, ..., $S_n$ determine if the relevant articles entail "Q" or "not Q". 

\[ Entails(S_1, S_2, ..., S_n , Q) \; or\]
\[ Entails(S_1, S_2, ..., S_n , \neg Q). \]

The answer of this task is binary: ``YES'' (``$Q$'') or ``NO'' (``$\neg Q$'').
The evaluation metric is accuracy and a random baseline (or simply giving always ``YES'' or always ``NO'') would lead to an accuracy about 0.5 since most test sets have an approximate equal distribution of positive and negative answers. 

One of the main challenges of this competition is the relatively small size of the training and test set.
Past competitions provide the answered questions of previous competitions resulting in a total of 806 questions.
The 2021 test set  contain 81 questions whereas the 2022 test set contained 109 questions.

%% file: previous_approaches.tex
\section{Prior work}

An overview of all COLIEE tasks and their respective approaches can be found in \citep{Rabelo_2022}. 
Most systems addressing task 4 were BERT-based \cite{devlin_bert:_2018} despite the small training set. Interestingly enough, the best performing system in 2021 utilized a method for increasing the training pool and deployed an ensemble of different BERT-based systems \cite{Yoshioka_2021}. 
Other systems relied on  different types on language models including DistilRoBERTa \cite{Liu:2019aa}, LEGAL-BERT \cite{chalkidis-etal-2020-legal}, T5 \cite{raffel-t5} and Electra \cite{Clark-electra} (i.e., \cite{schilder_pentapus_2021})

Other notable past systems used the Japanese original text and developed a rule-based system to identify entailment \cite{KIS} while another system was based on a Graph Neural network  using DistilRoBERTa and LEGAL-BERT embeddings as nodes \cite{OvGU}.

Only one system \cite{schilder_pentapus_2021} looked into a few-shot learning approach utilizing GPT-3 \cite{brown2020language}.
The system was based on a few-shot approach addressing  task 3 and task 4 together\footnote{The organizers offered a task 5 that would require a system to do task 3 (i.e., retrieval) and task 4 (i.e., entailment) in one step.}. 
Their experiments showed that GPT-3 without any further fine-tuning and without prompting of the relevant articles from the Japanese statues does not perform very well on this task reaching results even below the random baseline.
These results indicate that even large LMs have not stored enough knowledge about Japanese law and more advanced prompting and/or fine-tuning techniques are required.

Other prior work we draw from explores the utility of prompting and explanations in order to have LLM solve more complex tasks such as commonsense reasoning or mathematical word problems. 
We draw from recent work that proposed so-called Chain-of-Thought (CoT) prompts that have shown improved results of Zero-Shot (ZS) and Few-Shot (FS) approaches \cite{kojima2022large}. 

Previous approaches to incorporating explanations into the reasoning process also show improvements over standard prompting or fine-tuning approaches. 
By relying on the generation capability of LLMs to produce reasoning text, those approaches produce text that then is used to fine-tune the language model. 
In contrast to earlier approaches that relied on manually created explanations \cite{rajani-etal-2019-explain}, \citet{zelikman2022star} propose a method that automatically creates rationales using GPT-J for Arithmetic word problems, Commonsense QA, and Grad School Math problems. 
Their STaR system keeps the rationales if the LLM produces the correct answer and creates a new set of training data they train the LM with. 
They show significant improvements over previous approaches that are fine-tuned on the answers only and comparable performance to a much larger LM (i.e., GPT-3) on CommonsenseQA.

%% file: experiments.tex
 \section{Experiments and results}
 This section contains descriptions of all experiments we conducted using various zero and few-shot approaches, different prompting strategies and fine-tuning approaches with or without explanations.
  We applied approaches that have been shown to improve the SOTA performance for general common sense datasets to the COLIEE legal data. 
 We use GPT-3 as the LLM and test on the COLIEE 2021 and 2022 test sets. 
 All experiments were conducted with OpenAI's GPT-3 (text-davinci-002) model, the most competent GPT-3 engine of the four (davinci, curie, babbage and ada).
 We use accuracy as the evaluation metric since the COLIEE test sets have an approximate equal distribution of positive and negative answers. 
 
 \subsection{Zero-shot (ZS)} \label{zs}
We design and incorporate different prompts for the ZS setting, none of which involve domain specific information. 
In Section \ref{LRP}, we describe ZS with domain-specific prompts that integrate legal reasoning approaches. 

To ensure that the completion of GPT-3 is deterministic and the same each time it is executed with the same input, we set the temperature for all experiments to 0. 
Other GPT-3 parameters are set to their default values (Top P=1, Frequency penalty = 0, Presence penalty = 0). 
We use greedy decoding throughout this paper for simplicity.

For ZS, the input we give GPT-3 contains the following parts: the instructive prompts we designed, the premise-hypothesis pairs from the COLIEE dataset, and the phrase ``True or False?''. 
We experimented with three different prompts on GPT-3, as shown in Table \ref{tab:zs}. 
Simply adding the term ``following'' to specify the location of the premise can enhance GPT-3's accuracy from 0.7160 to 0.7407. 
If the prompt is less instructive, for instance, if ``the given premise'' is replaced with ``Japanese civil code statutes'', the accuracy decreases from 0.7407 to 0.7037. 
Note that ``the given premise'' consists of the Japanese civil code statutes that is most closely associated with the given hypothesis. 
Be aware that the accuracy 2021 COLIEE winner obtained was 0.7037, which is the same as our least performing prompt in the ZS setting. 
Thus, by simply providing a more specific, relevant, and instructive prompt to GPT-3, we can outperform the 2021 COLIEE winner by 3.70\% point of accuracy.

We use prompt2 (``Please determine if the following hypothesis is True or False based on the given premise.'') in all the following discussed approaches since it achieves the best accuracy in the ZS setting.

\input{tables/zs.tex}

\subsection{Few-shot (FS)}
\citet{brown2020language} proved that the FS performance of LLMs is superior to that of ZS and some SOTA fine-tuning techniques in a variety of tasks. 
We thus evaluate GPT-3 in the FS setting by giving GPT-3 examples of hypothesis-answer pairs from a blog\footnote{https://www.crear-ac.co.jp/shoshi/kakomon/} including previously evaluated bar exam questions. 
The original hypothesis-answer pairs from the blog are in Japanese, we use the Google-translated English version as the few shots.

\begin{figure}
\centering
  \includegraphics[width=1\linewidth]{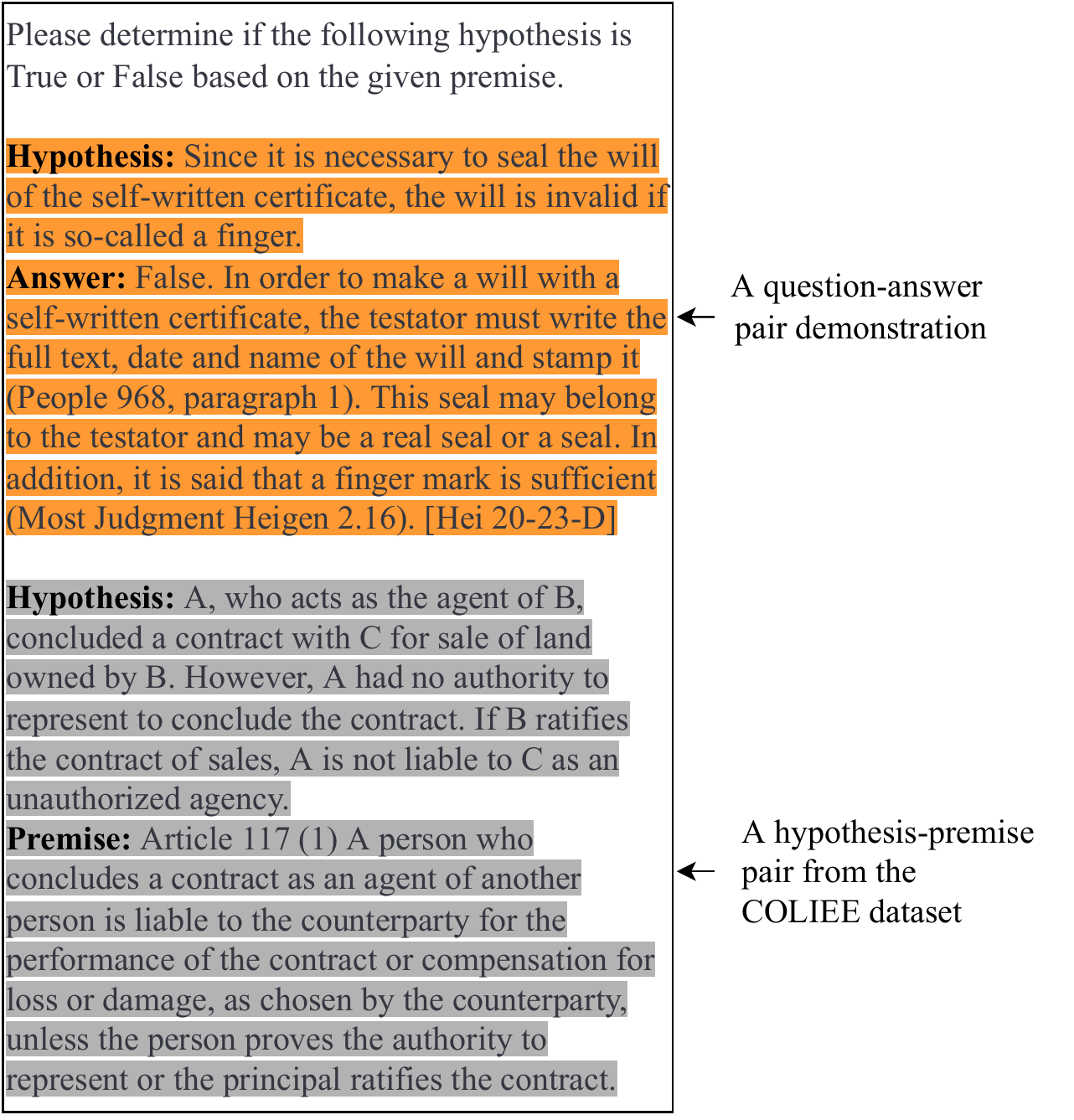}
  \caption{1-shot with a specific hypothesis-premise pair where the 1-shot is from a Japanese bar exam preparation blog and the hypothesis-premise pair is from the COLIEE 2021 test set.}
  \label{fig:1shot}
\end{figure}

We conducted experiments using 1-shot, 3-shot, and 8-shot, where one, three, and eight question-answer pairs are provided to GPT-3, respectively. 
The 1-shot example with a specific hypothesis-premise pair is shown in Figure \ref{fig:1shot}.

The result is shown in Table \ref{tab:FS}. 
On the 2021 COLIEE test set, 3-shot and 8-shot achieve the same accuracy, outperforming 1-shot, and all three exceed the 2021 COLIEE winner (0.7037). 
On the 2022 COLIEE test set, 8-shot outperforms 3-shot and 1-shot, and all three beat the 2022 COLIEE winner (0.6789).

\input{tables/fs.tex}

\subsection{Zero-shot-Chain of Thought (ZS-CoT)}
\begin{figure*}[!htb]
\centering
  \includegraphics[width=1\linewidth]{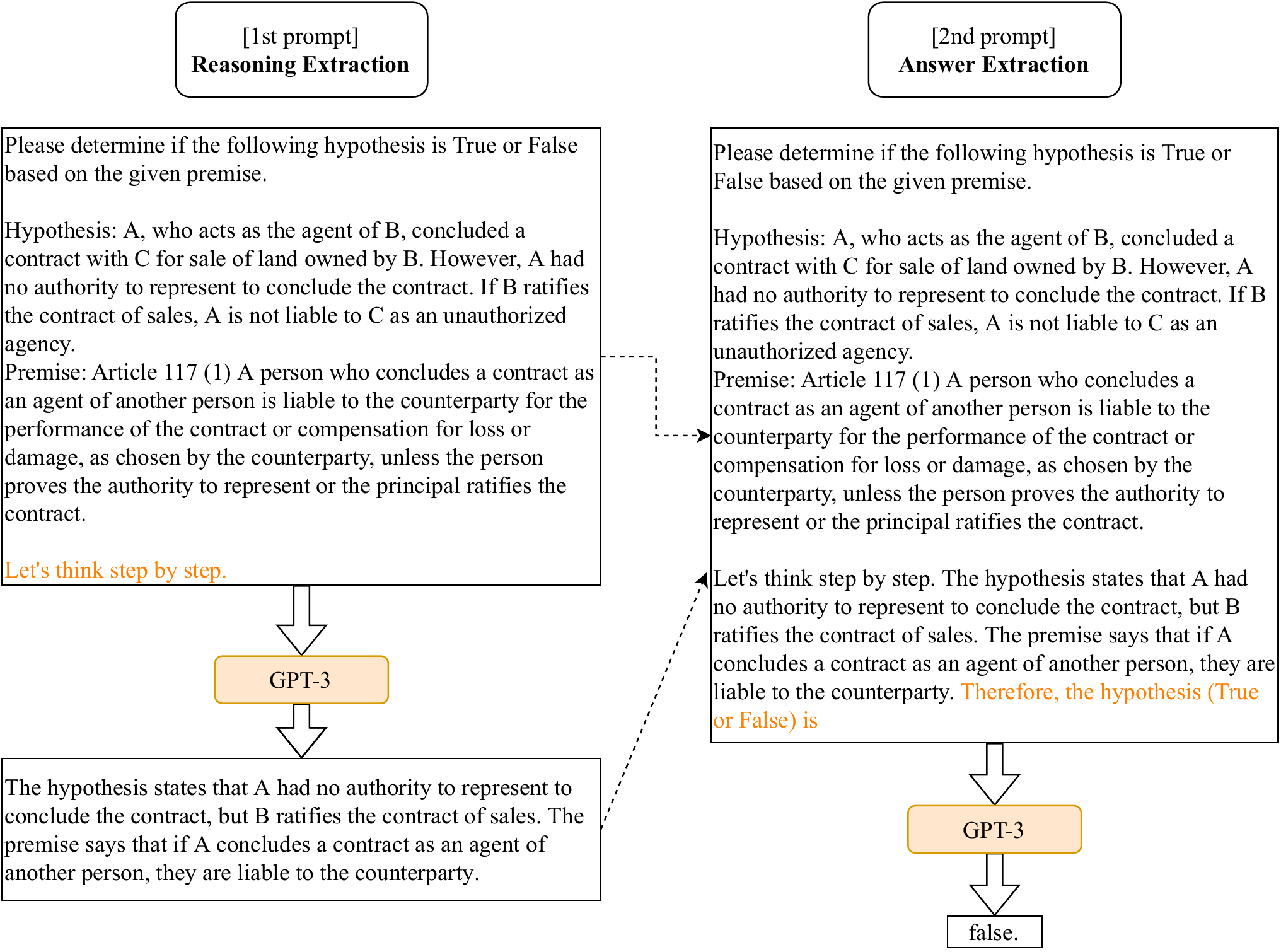}
  \caption{The complete ZS-CoT pipeline. First, we use the ``reasoning'' prompt to extract a complete reasoning process from GPT-3, then we use the ``answer'' prompt to extract the answer in the proper format from the reasoning text.}
  \label{fig:ZS-CoT}
\end{figure*}

\citet{kojima2022large} showed that LLMs can generate chain-of-thought reasoning processes and have superior reasoning capabilities by just adding ``Let's think step by step'' before each answer for common sense reasoning tasks. 
Hence, we experimented with GPT-3 using two-stage prompting (i.e., the output of the generated text from the first prompt is used for the second prompt): ``Let's think step by step'' and ``Therefore, the hypothesis is (True or False)''.
An illustration for the ZS-CoT pipeline using a hypothesis-premise pair from the 2021 COLIEE dataset is shown in Figure \ref{fig:ZS-CoT}.

Though ZS-CoT is theoretically straightforward, its nuance lies in the fact that it employs prompting twice. 
As shown in Figure \ref{fig:ZS-CoT}, in the first stage, which is the reasoning extraction process, we provide GPT-3 with \{prompt\} + \{premise\} + \{hypothesis\} + \{CoT\} as input, where the prompt is the prompt2 described in Section \ref{zs}, and CoT is ``Let's think step by step'' since this CoT prompt demonstrates to obtain the best performance in common sense reasoning tasks \cite{kojima2022large}. 
GPT-3 generates its reasoning process based on the input, but the generated completion may (or may not) contain the final answer (True or False in our case). 
In the second stage, which is the answer extraction process, we provide GPT-3 with both the input and output from the first stage as well as a second prompt (answer trigger): ``Therefore, the hypothesis (True or False) is''. 
The prompted text is passed to GPT-3 as input to create the final binary answer.

When ZS-CoT is applied to GPT-3 on the COLIEE test sets, however, the results are not as strong as the results reported by \citet{kojima2022large} for common sense datasets. 
The accuracies are 0.6296, and 0.7064 on the COLIEE 2021 and 2022 test sets, respectively. 
While ZS-CoT outperforms ZS on the 2022 test set, and outperforms the 2022 COLIEE winner by 4.05\%, it underperforms ZS on the 2021 test set by 17.65\%.

Notably, GPT-3 is capable of generating explanations in all of the previously stated approaches, including ZS, FS, and ZS-CoT, although not all ZS and FS completions contain explanations. 
Particularly when the predicted answer is True, explanations are often not provided. 
Explanations are always generated by GPT-3 when prompted by ZS-CoT though.

\subsection{Fine-tuning LM with and without explanations}
\input{tables/ft.tex}
Fine-tuning language models can usually gain strong performance on many benchmarks \cite{brown2020language}. 
It involves updating the weights of a pre-trained language model by training on an annotated dataset specific to the desired task. 
We fine-tune GPT-3 with the 2021 COLIEE training set and expect the fine-tuned model to achieve better performance compared to using the pre-trained model directly.

To fine-tune GPT-3, a collection of training samples consisting of the expected input and its corresponding output (``completion'') are required. 
We fine-tune GPT-3 using the 2021 COLIEE training set, where premise-hypothesis pairs are utilized in the input and answers are used in the completions. 
We fine-tune GPT-3 with two types of completions: binary answers only (True or False) and binary answers with explanations. 
The explanations are either explanations created by GPT-3 or pseudo-explanations extracted as stated below.

Although OpenAI's documentation states that no prompt is necessary in the input for fine-tuning GPT-3, we nevertheless utilize two forms of input: without prompt and with prompt2 specified in Section \ref{zs} to examine the effect of prompts during GPT-3's fine-tuning process. 
The fine-tuned model is then tested on the 2021 COLIEE test set. 
The result is shown in Table \ref{tab:fine-tune}. 
It is shown that GPT-3 fine-tuned with instructive prompts outperforms the setting without prompts by 23.99\%.

When fine-tuning GPT-3 with both binary answers and explanations, we use two types of explanations that are created using different approaches.

\begin{figure}
\centering
  \includegraphics[width=0.8\linewidth]{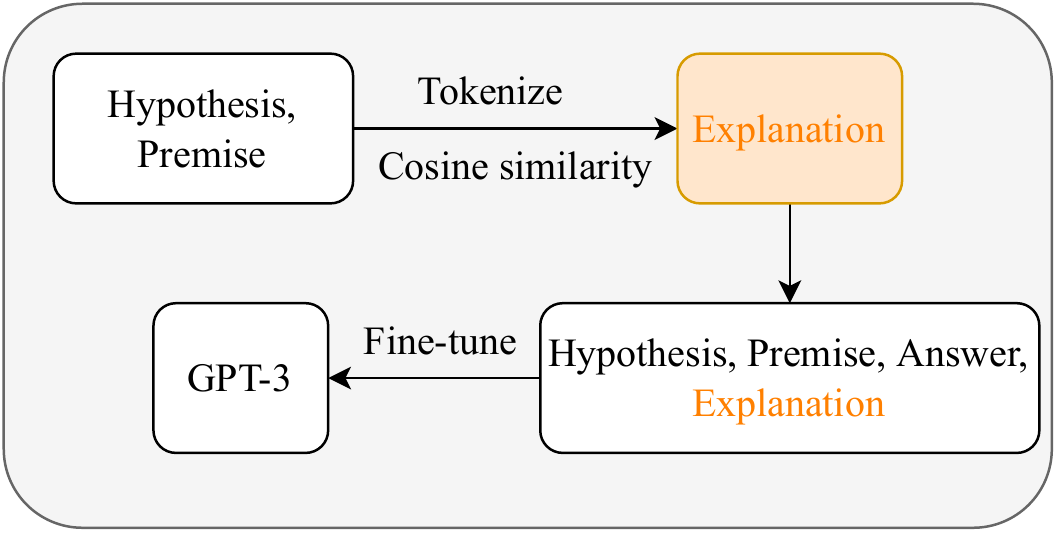}
  \caption{The pipeline of fine-tuning GPT-3 with pseudo-explanation.}
  \label{fig:ft-pseudo}
\end{figure}

\textbf{Pseudo-explanation.} 
We select from each premise the sentence that is most relevant to its corresponding hypothesis as the pseudo-explanation. 
More specifically, for each hypothesis-premise pair, we first split the premise into sentences, encode each sentence using the MPNet encoder \cite{song2020mpnet}, compute the cosine similarity score between each sentence and the hypothesis, then choose the sentence with the highest similarity score as the explanation. 
Since fine-tuning with binary answers achieves higher accuracy when an instructive prompt is included in the input, hence we provide GPT-3 with \{prompt\} + \{premise\} + \{hypothesis\} + ``True or False'' as input and \{label\} + ``Because according to '' + \{pseudo-explanation\} as completion during the fine-tuning process with both binary answers and explanations. 
In this manner, we assist GPT-3 in determining which portion of the premise it should focus on in order to arrive at the final response. 
Similar to the completion used to fine-tune GPT-3, the completion generated during inference also contains binary answer and explanation, which is one sentence from the premise with the greatest similarity score to the hypothesis. 
The fine-tuning process is illustrated in Figure \ref{fig:ft-pseudo}, and the inference results can be seen in Table \ref{tab:fine-tune}.

\begin{figure}
\centering
  \includegraphics[width=0.8\linewidth]{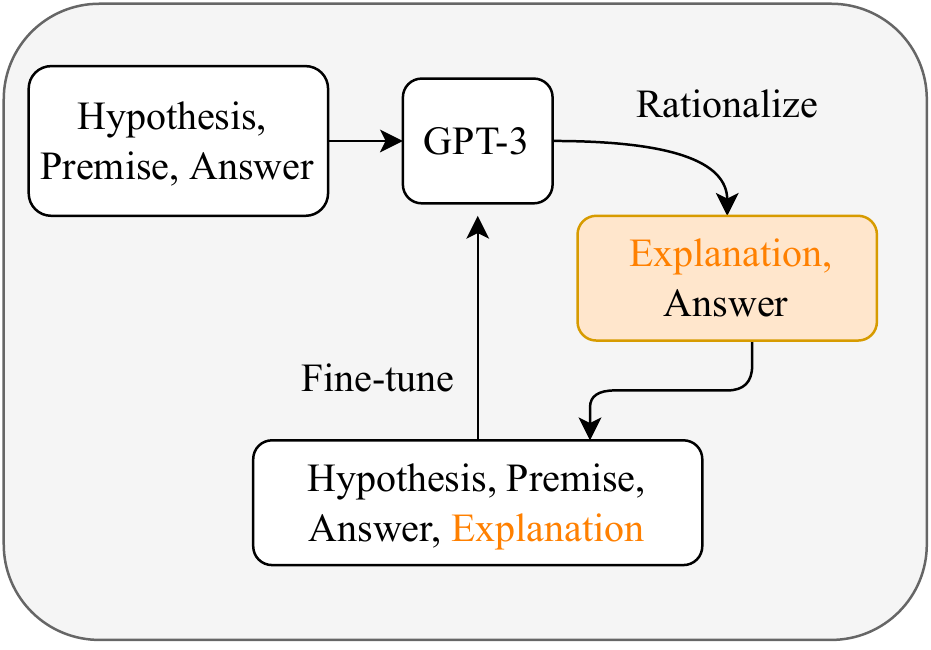}
  \caption{The pipeline of fine-tuning GPT-3 with GPT-3-generated-explanation.}
  \label{fig:STaR}
\end{figure}

\textbf{GPT-3-generated explanation.} 
This approach is inspired by the ``Self-Taught Reasoner'' (STaR bootstrapping) approach \cite{zelikman2022star}, which iteratively employs LLM-generated explanations to bootstrap the LLM's capacity to execute more complicated reasoning. 
In a single loop of STaR bootstrapping: LLM creates explanations to answer a given question; if the answer generated is incorrect, LLM will retry to generate an explanation based on the right answer and then fine-tune on the explanations that produce correct answers. 
Since repeatedly fine-tuning GPT-3 is a costly process, we provide an alternative to the STaR bootstrapping approach, which is shown in Figure \ref{fig:STaR}. 
In our approach, we prompt GPT-3 to generate explanations for each hypothesis-premise-answer triplets by providing it with ``Please explain why the following hypothesis is'' + \{label\} + ``based on the given premise.'' + \{premise\} + \{hypothesis\}, then use the hypothesis-premise-answer-explanation quartets to fine-tune GPT-3, that is, providing GPT-3 with input: \{prompt\} + \{premise\} + \{hypothesis\} + ``True or False'' , and completion: \{explanation\}, where the explanation is the GPT-3-generated explanations from the previous step.
To decrease cost, we only fine-tune GPT-3 once as opposed to repeatedly. 
During inference, the fine-tuned model could provide both binary answers and plausible-seeming explanations. 
The inference accuracy is shown in Table \ref{tab:fine-tune}.

As shown in Table \ref{tab:fine-tune}, fine-tuning GPT-3 with pseudo-explanation surpasses fine-tuning GPT-3 with its own explanation. 
One possible reason is that GPT-3-generated explanations are not always accurate, and fine-tuning with incorrect explanations will decrease accuracy.
Furthermore, fine-tuning GPT-3 with either pseudo-explanation or GPT-3-generated explanation underperforms fine-tuning GPT-3 with instructive prompts and labels alone. 
This indicates that encouraging GPT-3 to ``think'' and ``reason'' independently by providing it with instructive prompts and less intervention could lead to better results.

\subsection{Legal Reasoning (LR) prompt} \label{LRP}
A common assumption in the literature is that prompts function as semantically meaningful task instructions and it requires experts to precisely define the task at hand \cite{webson-pavlick-2022-prompt}. 
Reformatting NLP tasks with varied prompts significantly increased ZS and FS performance compared to traditional fine-tuned models \cite{wei2021finetuned,le-scao-rush-2021-many,schick-schutze-2021-just,sanh2021multitask}. 
For COLIEE task 4, the above discussed experiments show that fine-tuning GPT-3 and FS with 3 or 8 examples outperform ZS. 
We thus utilize legal reasoning prompts to boost GPT-3's accuracy under the ZS setting.

\input{tables/legalprompt.tex}

``Legal reasoning'', ``creative thinking'' and ``critical analysis'' are the pillars to ``think like a lawyer'' and contribute to the formation of a professional legal identity \cite{kift2011bachelor}. 
\citet{burton2017think} identified a number of acronyms used to teach traditional legal reasoning, which appear to represent legal reasoning approaches often used by legal experts and practitioners. 
For example, approach IRAC is the acronym for \textit{Issue, Rule, Application, Conclusion}. 
Here, \textit{issue} means issue spotting, or thinking about what facts and circumstances brought the parties to court. 
\textit{Rule} means to find the governing law for the issue. 
\textit{Application} refers to the process of applying the rule to the facts of the issue.
\textit{Conclusion} is to come up with the conclusion from the application as to whether the rule applies to the facts. 
\begin{figure}
\centering
  \includegraphics[width=1\linewidth]{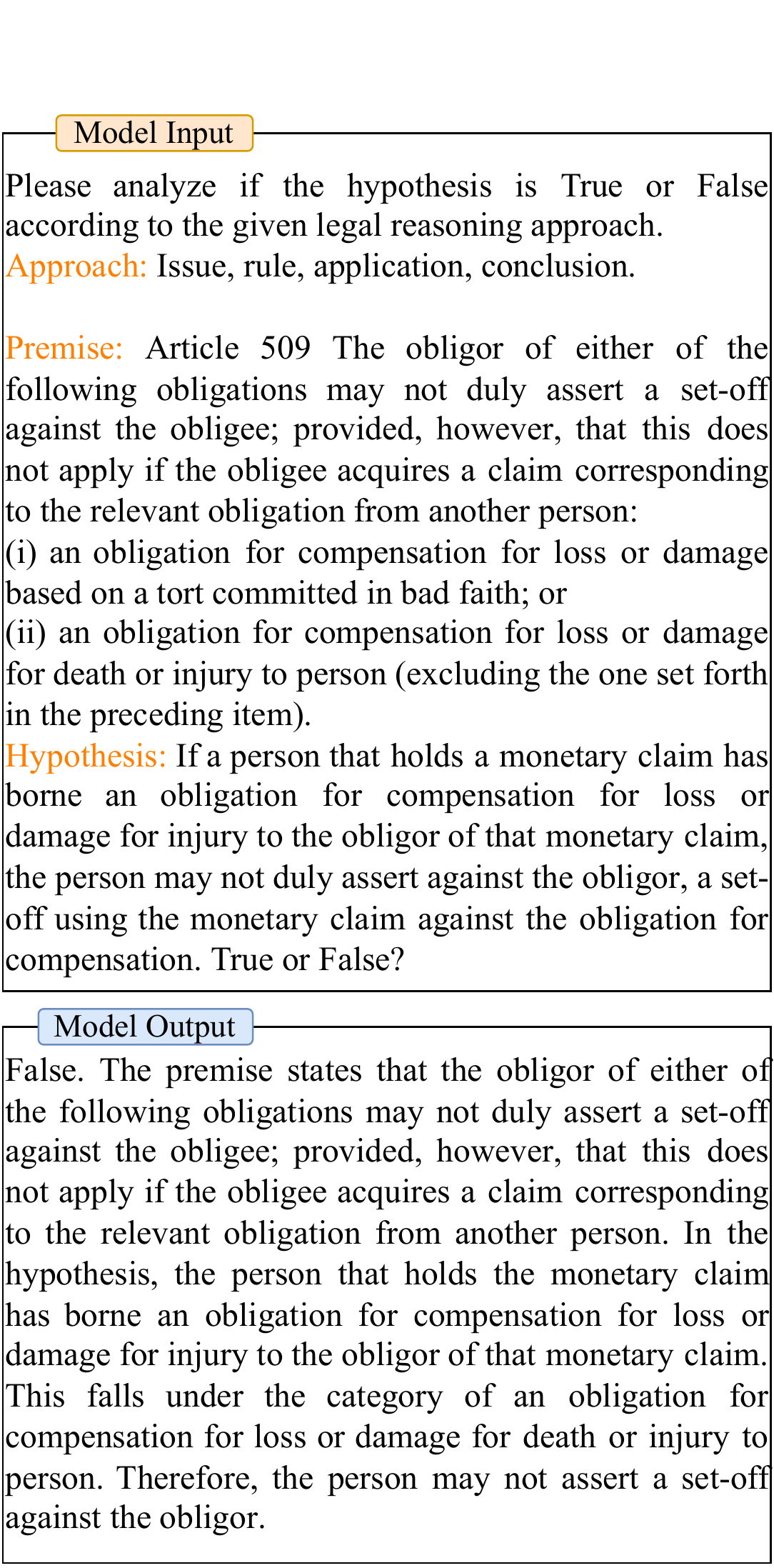}
  \caption{GPT-3's input and output when applying ZS with a LR prompt. Here, the LR approach used is IRAC.}
  \label{fig:ZS-LR}
\end{figure}
We thus ask GPT-3 to think like a lawyer by prompting it with \{prompt\} + \{approach\} + \{premise\} + \{hypothesis\} + ``True or False?''. 
We use ``Please analyze if the hypothesis is True or False according to the given legal reasoning approach'' as the prompt.
An illustration of GPT-3's input and output when applying the LR prompt in the ZS setting is shown in Figure \ref{fig:ZS-LR}.
The approach is extracted from the legal reasoning approaches summarized by \citet{burton2017think}, which can be found in Table \ref{tab:legalprompt}. 
As indicated in the table, TRRAC beats COLIEE winner by 15.79\% in 2021, and IRREAC and IRRAC beat COLIEE winner by 5.41\% in 2022. 
Note also that GPT-3 could provide explanations that seem to adhere to the provided legal reasoning method. Some explanation examples are shown in the Appendix.

Figure \ref{fig:acc_comparison} shows a comparison of the accuracy of each approach applied to GPT-3. For the ZS approach we select the best-performing schema that is when prompt2 is used; for ZS with legal reasoning prompts approach, we select the best-performing legal reasoning approach for 2021 and 2022, which are TRRAC, and IRREAC/IRRAC. 
For the fine-tuning approach, we fine-tune and test GPT-3 on the same year's training and test set provided by the COLIEE organizer. 

\begin{figure}
\centering
  \includegraphics[width=1\linewidth]{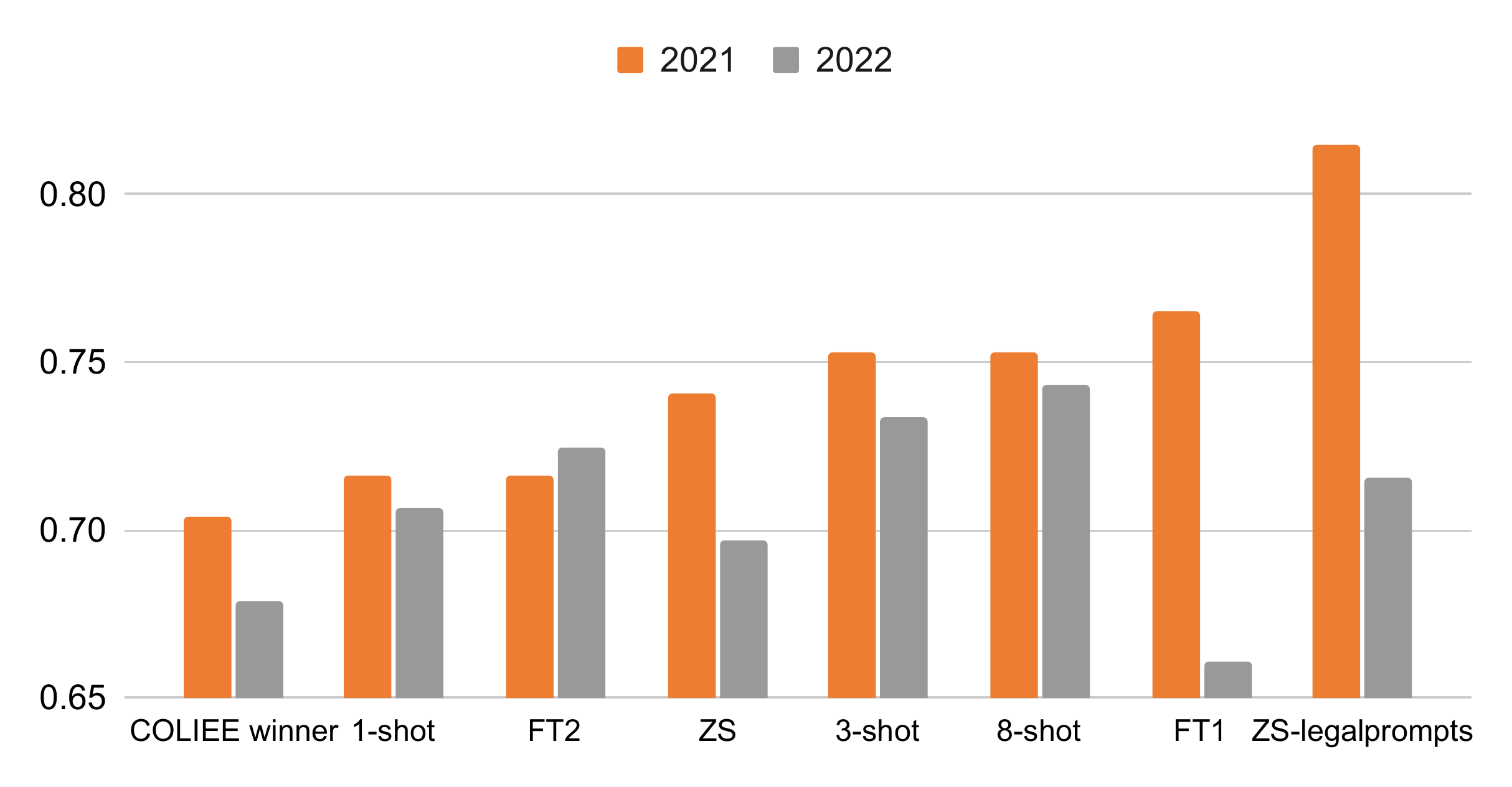}
  \caption{Comparison of the accuracy of COLIEE winners and GPT-3 when different approaches are used. FT1 refers to fine-tuning GPT-3 with binary answers and prompt, FT2 refers to fine-tuning GPT-3 with pseudo-explanation.}
  \label{fig:acc_comparison}
\end{figure}

%% file: tables/zs.tex
\begin{table*}[!htb]
\centering
\caption{GPT-3's ZS performance with different prompts on the COLIEE 2021 test set. Minor changes in the prompt can impact GPT-3's accuracy largely. }
\label{tab:zs}
\resizebox{\textwidth}{!}{%
\begin{tabular}{|c|c|c|}
\hline
\textbf{Input}                                                                                                             & \textbf{Prompt}                                                                                                                                              & \textbf{Accuracy} \\ \hline
\multirow{3}{*}{\begin{tabular}[c]{@{}c@{}}\{prompt\} + \{premise\} + \\ \{hypothesis\} + ``True or False?''\end{tabular}} & Prompt1: Please determine if the hypothesis is True or False based on the given premise.                                                                     & 0.7160            \\ \cline{2-3} 
                                                                                                                           & \begin{tabular}[c]{@{}c@{}}Prompt2: Please determine if the following hypothesis is True or False based on the \\ given premise.\end{tabular}                & 0.7407            \\ \cline{2-3} 
                                                                                                                           & \begin{tabular}[c]{@{}c@{}}Prompt3: Please determine if the following hypothesis is True or False based on the \\ Japanese civil code statutes.\end{tabular} & 0.7037            \\ \hline
\end{tabular}
}
\end{table*}

%% file: tables/fs.tex
\begin{table}[!htb]
\centering
\caption{GPT-3's accuracy when provided with different numbers of examples as demonstrations. All three settings outperform 2021 and 2022 COLIEE winners (0.7037 and 0.6789, respectively).}
\label{tab:FS}
\begin{tabular}{|c|c|c|}
\hline
\textbf{Number of shots} & \textbf{2021} & \textbf{2022} \\ \hline
1-shot                   & 0.7160        & 0.7064        \\ \hline
3-shot                   & 0.7531        & 0.7339        \\ \hline
8-shot                   & 0.7531        & 0.7432        \\ \hline
\end{tabular}
\end{table}

%% file: tables/ft.tex
\begin{table*}[b]
\centering
\caption{The accuracy achieved by the fine-tuned GPT-3 on the 2021 COLIEE test set. GPT-3 were fine-tuned with various input and completion. The prompt in the input is ``Please determine if the following hypothesis is True or False based on the given premise.''}
\label{tab:fine-tune}
\resizebox{\textwidth}{!}{%
\begin{tabular}{|c|c|c|}
\hline
\textbf{Input during Fine-tuning}                            & \textbf{Completion during Fine-tuning}                       & \multicolumn{1}{l|}{\textbf{Accuracy}} \\ \hline
\{premise\} + \{hypothesis\} + ``True or False?''              & \{label\}                                                    &    0.6173                                    \\ \hline
\{prompt\} + \{premise\} + \{hypothesis\} + ``True or False?''& \{label\}                                                    & 0.7654                                 \\ \hline
\{prompt\} + \{premise\} + \{hypothesis\} + ``True or False?'' & \{label\} + ``Because according to '' + \{pseudo-explanation\} & 0.7160                                 \\ \hline
\{prompt\} + \{premise\} + \{hypothesis\} + ``True or False?'' & \{GPT-3-generated explanation\}                              &  0.6173                                      \\ \hline
\end{tabular}%
}
\end{table*}

%% file: tables/legalprompt.tex
\begin{table*}[!htb]
\centering
\caption{GPT-3's performance on the 2021 and 2022 COLIEE test sets by applying various legal reasoning approaches to the ZS setting. The best performing approach in 2021 TRRAC outperforms the 2021 COLIEE winner by 15.79\%, and the best performing approaches in 2022, IRREAC and IRRAC, surpass the 2022 COLIEE winner by 5.41\%.}
\label{tab:legalprompt}
\resizebox{\textwidth}{!}{%
\begin{tabular}{|c|c|c|c|}
\hline
\textbf{Approach} & \textbf{Details}                                                                                                                                                                                                                                                                                         & \textbf{Accuracy (2021)} & \textbf{Accuracy(2022)} \\ \hline
\textbf{TRRAC}             & \textbf{Thesis, rule, rule, application, conclusion}                                                                                                                                                                                                                                                              & \textbf{0.8148}                   & 0.6881                  \\ \hline
CLEO              & Claim, law, evaluation, outcome                                                                                                                                                                                                                                                                          & 0.8025                   & 0.6881                  \\ \hline
ILAC              & Issue, law, application, conclusion                                                                                                                                                                                                                                                                      & 0.7778                   & 0.6972                  \\ \hline
IRAACP            & Issue, rule, apply, apply, conclusion, policy                                                                                                                                                                                                                                                            & 0.7778                   & 0.6881                  \\ \hline
\textbf{IRREAC}            & \textbf{Issue, rule, rule, application, conclusion}                                                                                                                                                                                                                                                               & 0.7778                   & \textbf{0.7156}                  \\ \hline
IGPAC             & Issue, general rule, precedent, application, conclusion                                                                                                                                                                                                                                                  & 0.7654                   & 0.6697                  \\ \hline
IPAAC             & Issue, principle, authority, application, conclusion                                                                                                                                                                                                                                                     & 0.7654                   & 0.6606                  \\ \hline
\textbf{IRRAC}             & \textbf{Issue, rule, reasoning, application, conclusion}                                                                                                                                                                                                                                                          & 0.7654                   & \textbf{0.7156}                  \\ \hline
IRAC            & Issue, rule, application, conclusion                                                                                                                                                                                                                                                      &       0.7284            &          0.6881     \\ \hline
\end{tabular}%
}
\end{table*}

%% file: conclusions.tex
\section{Conclusions and discussion}
Over the past several years, significant focus has been placed on leveraging reasoning-based-prompt approaches for increasing system performance in large language model-based question answering (e.g., \cite{rajani-etal-2019-explain,kojima2022large,zelikman2022star}). 
While simple zero or few-shot approaches are well-performant on basic benchmark or trivia tasks, elaborate reasoning-based strategies are required for more complex domains, such as scientific examinations or legal entailment, where successfully answering a query requires following a chain of logic from a stated (or implicit) context.
Practitioners specifically within the legal domain will often follow certain well-defined approaches to reasoning, such as \textit{Issue, Rule, Application, Conclusion}, and our hypothesis was that prompting language models in a similar fashion -- alongside explanation-based fine-tuning -- could improve capabilities against complex reasoning queries.

Leveraging data from the 2021 and 2022 COLIEE competitions \cite{Rabelo_2022} for training/testing, our exploration found that the most significant improvements in accuracy can be achieved by applying zero-shot queries to GPT-3 legal-reasoning (i.e., \textit{Thesis, rule, rule, application, conclusion} \cite{burton2017think}) prompts on 2021 data, but these approaches did not perform that well on the 2022 test data.
 A 8-shot few-shot learning approach turned out to show the best results for the 2022 data. 
Nevertheless, both approaches significantly outperform COLIEE state-of-the-art, improving the 2021 best system from 0.7037 accuracy to 0.8148, and the 2022 best system from 0.6789 accuracy to 0.7431. 

Additionally, we like to emphasize certain counterintuitive experimental findings. In our experiment, 1-shot underperform zero-shot by 2.47\% point of accuracy on 2021 test set, and fine-tuning performed worse than zero-shot when applying the same prompt.
We argue that zero-shot is our baseline and reflects GPT-3's intrinsic reasoning capability, and the question-answer pair example we used in 1-shot may be irrelevant to any of the questions in the test set, which may have misled GPT-3 and reduced its performance. 
Therefore, an idea for future work is to carefully select the few-shot examples, ensuring that the topics covered in the test sets are the same as or relevant to those in the examples.

Further, we find that several other approaches also perform comparatively well versus COLIEE competition winners, such as zero shot chain-of-thought \cite{zelikman2022star} outperforming the 2022 COLIEE winner by 4.05\% (but underperforming the 2021 results by 17.65\%) and fine-tuning with pseudo-explanations explicitly derived from the query data's premise (yielding 71.6\% accuracy on the 2021 test set). The few-shot approaches show the most promising because they achieved good and consistent performance results over the two years, while the zero-shot with legal reasoning prompts show outstanding results for one year but lower results for the other year. This indicates that such prompting approaches require to be tested more thoroughly and need to be investigated further. 
Although the performance of zero-shot with legal reasoning prompts is inconsistent between the two years, for instance, TRRAC had the highest accuracy in 2021 but the third highest accuracy in 2022, all legal reasoning approaches we tested outperformed COLIEE winners for 2021, and half outperformed COLIEE winners for 2022.
Future study could construct an ensemble method for selecting the final binary answer for each question by referencing the responses obtained from multiple legal reasoning approaches based on majority voting.

While our analysis shows significant promise in prompt engineering for high-order LLM-based reasoning tasks, it is questionable whether prompting actually teaches a LM to ``think like a lawyer'' and further exploration is warranted to determine the likelihood of success of fine-tuning with explanation and (legal) reasoning prompting on the legal or other domains and data sets.

%% file: appendix.tex
Some explanation examples generated by GPT-3 using ZS with the TRRAC legal reasoning prompt on 2021 COLIEE test set are shown below.

\textbf{ID}: R02-24-O

\textbf{Hypothesis}: In the case that the seller cannot acquire and transfer the rights to the buyer, the buyer may not make a claim for damages to the seller in good faith without giving a notice within a year from the time when the seller knows the fact that the rights belong to others.

\textbf{Premise}: Article 566 If the subject matter delivered by the seller to the buyer does not conform to the terms of the contract with respect to the kind or quality, and the buyer fails to notify the seller of the non-conformity within one year from the time when the buyer becomes aware of it, the buyer may not demand cure of the non-conformity of performance, demand a reduction of the price, claim compensation for loss or damage, or cancel the contract, on the grounds of the non-conformity; provided, however, that this does not apply if the seller knew or did not know due to gross negligence the non-conformity at the time of the delivery.

\textbf{Answer and explanation}: False. 
The buyer may not make a claim for damages to the seller in good faith without giving a notice within a year from the time when the seller knows the fact that the rights belong to others, but the buyer may demand cure of the non-conformity of performance, demand a reduction of the price, claim compensation for loss or damage, or cancel the contract.

\vspace{5mm} 
\textbf{ID}: R02-27-O

\textbf{Hypothesis}: A partner may seek the division of the partnership property before liquidation if the consent of the majority of partners exists.

\textbf{Premise}: Article 676 (3) A partner may not seek the division of the partnership property before liquidation.

\textbf{Answer and explanation}: False. 
The rule states that a partner may not seek the division of the partnership property before liquidation. 
The application of the rule is that the consent of the majority of partners does not exist. 
Therefore, the conclusion is that a partner may not seek the division of the partnership property before liquidation.

\vspace{5mm} 
\textbf{ID}: R02-12-A

\textbf{Hypothesis}: An obligor (A) created a mortgage for an obligee (B) on the real property owned by A and the creation was registered. 
A may assert the extinctive prescription of the mortgage if 20 years have passed from the time when it becomes exercisable, even if the claim the mortgage secures is not extinguished.

\textbf{Premise}: Article 396 No mortgage is extinguished by prescription in relation to obligors and mortgagors unless it is extinguished simultaneously with the claim the mortgage secures.

\textbf{Answer and explanation}: False. 
The mortgage is not extinguished by prescription unless the claim it secures is extinguished.

\vspace{5mm} 
\textbf{ID}: R02-1-A

\textbf{Hypothesis}: The family court may decide to commence an assistance also in respect of a person whose capacity to appreciate their own situation is extremely inadequate due to a mental disorder.

\textbf{Premise}: Article 15 (1) The family court may decide to commence an assistance in respect of a person whose capacity to appreciate their own situation is inadequate due to a mental disorder, at the request of the person in question, that person's spouse, that person's relative within the fourth degree of kinship, the guardian, the guardian's supervisor, the curator, the curator's supervisor, or a public prosecutor; provided, however, that this does not apply to a person with respect to whom there are grounds as prescribed in Article 7 or the main clause of Article 11.
Article 11 The family court may decide to commence a curatorship in respect of a person whose capacity to appreciate their own situation is extremely inadequate due to a mental disorder, at the request of the person in question, the person's spouse, the person's relative within the fourth degree of kinship, the guardian, the guardian's supervisor, the assistant, the assistant's supervisor, or a public prosecutor; provided, however, that this does not apply to a person in respect of whom a cause set forth in Article 7 exists.

\textbf{Answer and explanation}: False. 
The family court may only decide to commence an assistance in respect of a person whose capacity to appreciate their own situation is inadequate due to a mental disorder, not a person whose capacity to appreciate their own situation is extremely inadequate due to a mental disorder.

\vspace{5mm} 
\textbf{ID}: R02-4-A

\textbf{Hypothesis}: A, who acts as the agent of B, concluded a contract with C for sale of land owned by B. 
However, A had no authority to represent to conclude the contract. 
A is liable for the performance of the contract or compensation for loss or damage, as chosen by A.

\textbf{Premise}: Article 117 (1) A person who concludes a contract as an agent of another person is liable to the counterparty for the performance of the contract or compensation for loss or damage, as chosen by the counterparty, unless the person proves the authority to represent or the principal ratifies the contract.

\textbf{Answer and explanation}: False. 
A is not liable for the performance of the contract or compensation for loss or damage, as chosen by A.

\vspace{5mm} 
\textbf{ID}: R02-8-E

\textbf{Hypothesis}: A took the jewelry that B had forgotten, believing without negligence that it belonged to A. 
In this case, A may not obtain the ownership of the jewelry by good faith acquisition.

\textbf{Premise}: Article 192 A person that commences the possession of movables peacefully and openly by a transactional act acquires the rights that are exercised with respect to the movables immediately if the person possesses it in good faith and without negligence.

\textbf{Answer and explanation}: The hypothesis is true. 
A cannot obtain the ownership of the jewelry by good faith acquisition because B did not give A permission to take the jewelry and A did not have knowledge that the jewelry belonged to B.

\vspace{5mm} 
\textbf{ID}: R02-25-E

\textbf{Hypothesis}: If the lessee lawfully subleases a leased thing, the lessor may not duly assert against the sublessee the cancellation by agreement of the lease with the lessee even if the lessor has a right to cancel due to non-performance on the part of the lessee at the time of the cancellation.

\textbf{Premise}: Article 613 (3) If the lessee lawfully subleases a leased thing, the lessor may not duly assert against the sublessee the cancellation by agreement of the lease with the lessee; provided, however, that this does not apply if, at the time of the cancellation, the lessor has a right to cancel due to non-performance on the part of the lessee.

\textbf{Answer and explanation}: False. 
The lessor may assert the cancellation by agreement of the lease against the sublessee if the lessor has a right to cancel due to non-performance on the part of the lessee at the time of the cancellation.

\vspace{5mm} 
\textbf{ID}: R02-1-U

\textbf{Hypothesis}: If the grounds of commencement of assistance cease to exist, the family court may rescind the decision for commencement of assistance without any party's request.

\textbf{Premise}: Article 18 (1) If the grounds prescribed in the main clause of Article 15, paragraph (1) cease to exist, the family court must rescind the decision for commencement of assistance at the request of the person in question, that person's spouse, that person's relative within the fourth degree of kinship, the guardian of a minor, the supervisor of a minor's guardian, the assistant, the assistant's supervisor, or a public prosecutor.
(2) At the request of a person as prescribed in the preceding paragraph, the family court may rescind all or part of the decision referred to in paragraph (1) of the preceding Article.

\textbf{Answer and explanation}: False. The family court may rescind the decision for commencement of assistance only at the request of the person in question, that person's spouse, that person's relative within the fourth degree of kinship, the guardian of a minor, the supervisor of a minor's guardian, the assistant, the assistant's supervisor, or a public prosecutor.

\vspace{5mm} 
\textbf{ID}: R02-26-I

\textbf{Hypothesis}: The provisions regarding mandate of juridical acts do not apply mutatis mutandis to entrustments of business that do not constitute juridical acts.

\textbf{Premise}: Article 656 The provisions of this Section apply mutatis mutandis to entrustments of business that do not constitute juridical acts.

\textbf{Answer and explanation}: False. 
The provisions of Article 656 do apply mutatis mutandis to entrustments of business that do not constitute juridical acts.